\pdfoutput=1

\documentclass[11pt]{article}

\usepackage[final]{acl}

\usepackage{times}
\usepackage{latexsym}

\usepackage[T1]{fontenc}

\usepackage[utf8]{inputenc}

\usepackage{microtype}

\usepackage{inconsolata}

\usepackage{graphicx}
\usepackage{hyperref}       
\usepackage{url}            
\usepackage{booktabs}       
\usepackage{amsfonts}       
\usepackage{nicefrac}       
\usepackage{xcolor}         

\usepackage{amsmath}
\usepackage{multirow}
\usepackage{colortbl}
\definecolor{Color1}{RGB}{210, 170, 230}
\usepackage{enumitem}
\usepackage{pifont}

\usepackage[ruled, lined, commentsnumbered, longend]{algorithm2e}

\newcommand*\samethanks[1][\value{footnote}]{\footnotemark[#1]}

\title{Rethinking Token Reduction for State Space Models}

\author{
Zheng Zhan$^1$\thanks{Equal contribution.},
Yushu Wu$^1$\samethanks[1],
Zhenglun Kong$^{12}$\samethanks[1],
Changdi Yang$^1$, \\
\textbf{Yifan Gong}$^1$,
\textbf{Xuan Shen}$^1$,
\textbf{Xue Lin}$^1$,
\textbf{Pu Zhao}$^1$,
\textbf{Yanzhi Wang}$^1$ \\
  $^1$Northeastern University, $^2$Harvard University \\
  \texttt{\{zhan.zhe, wu.yushu, p.zhao, yanz.wang\}@northeastern.edu} 
}

\begin{document}
\maketitle
\begin{abstract}
Recent advancements in State Space Models (SSMs) have attracted significant interest, particularly in models optimized for parallel training and handling long-range dependencies. Architectures like Mamba have scaled to billions of parameters with selective SSM. To facilitate broader applications using Mamba, exploring its efficiency is crucial. While token reduction techniques offer a straightforward post-training strategy, we find that applying existing methods directly to SSMs leads to substantial performance drops. Through insightful analysis, we identify the reasons for this failure and the limitations of current techniques. In response, we propose a tailored, unified post-training token reduction method for SSMs. Our approach integrates token importance and similarity, thus taking advantage of both pruning and merging, to devise a fine-grained intra-layer token reduction strategy. Extensive experiments show that our method improves the average accuracy by 5.7\% to 13.1\% on six benchmarks with Mamba-2 compared to existing methods, while significantly reducing computational demands and memory requirements.\footnote{Code available at \url{https://github.com/wuyushuwys/ToR_SSM}}
\end{abstract}

\section{Introduction}

There are growing research interests and efforts in SSMs in recent years.
Building on the foundation laid by the Kalman filter model~\cite{kalman1960new}, SSMs have evolved to address long-range dependencies and are optimized for parallel training. Several works~\cite{gu2021efficiently, gu2021combining, gu2022parameterization, gupta2022diagonal, dao2024transformers} have proposed SSM-based models capable of processing sequence data across a variety of tasks and modalities.

A notable recent contribution, Mamba~\cite{originalmamba}, integrates time-varying parameters into SSMs, allowing the model to selectively propagate or forget information. Additionally, Mamba introduces a hardware-aware parallel algorithm that accelerates both training and inference. Unlike quadratic attention mechanisms, which become prohibitively expensive with longer sequence lengths, Mamba's subquadratic-time architecture is more efficient and better suited for handling long sequences.
The exceptional scaling performance of Mamba underscores its potential as an effective alternative to the Transformer model~\cite{vaswani2017attention} for generative language modeling tasks.

In line with existing research efforts aimed at enhancing the efficiency of Transformer models \cite{shen2024edgeqat,shen2024search,Zhan_2021_ICCV}, exploring the efficiency of SSMs is crucial for facilitating real-time applications. While weight pruning and quantization are prevalent techniques for optimizing Transformer models \cite{vaswani2017attention,yang2023pruning,zhang2022advancing}, token reduction \cite{rao2021dynamicvit,pan2021ia,yuan2021tokens,renggli2022learning} has proven effective in improving Transformer efficiency due to 
the token length dimension or number of token is independent of the model architecture.  

Given that SSM blocks also process input tokens similarly to Transformer models, applying existing state-of-the-art (SOTA) token reduction techniques \cite{liang2022evit,cao2023pumer,bolya2022tome} to SSMs appears to be a straightforward post-training approach to enhance their efficiency, especially when scaling to billions of model parameters. This can achieve faster serving and lower peak memory usage, facilitating the wider deployment of large-scale SSMs like Mamba. However, as illustrated in Figure \ref{fig:previous}, this application of token reduction to SSMs, while offering some benefits of faster inference with fewer tokens, results in significant performance drops.

In this paper, after  applying  existing Transformer token reduction techniques  to SSMs and observing their failures, 
we conduct an insightful analysis to understand the patterns and reasons for their failures on SSMs.
Based on our analysis, we propose a unified post-training token reduction method for SSMs to preserve performance and improve efficiency. We first employ a decoupling strategy that computes the importance of each token and classifies them into two sets: less important tokens and more important tokens. Following this, we devise a fine-grained intra-layer token reduction strategy for the hidden states and residual connections of Mamba. 
Our approach uses a hybrid token reduction strategy (combining and taking advantages of pruning and merging) on hidden state tokens, meticulously designed to balance preserving essential information and eliminating redundancy. Our unified strategy can be generalized to other model architectures like Transformers.
In summary, the main contributions of our work are as follows:

\begin{itemize}[leftmargin=*]
    \item We observe the failure of directly applying token reduction techniques from Transformers to SSMs, and we conduct an insightful analysis to investigate the patterns of token reduction strategies and the possible reasons for their failures.
    \item We are the first to propose a unified post-training token reduction method designed for SSMs. This strategy leverages insights from both token pruning and token merging, and incorporates the token importance and similarity evaluation.  
    \item Zero-shot evaluations on various SSMs demonstrate the effectiveness of our method, improving average accuracy by 5.7\% to 13.1\% on six benchmarks with Mamba-2, and by 6.5\% to 15.1\% with Mamba compared to baseline methods. Meanwhile, our method significantly reduces computational demands and memory requirements.
\end{itemize}
\section{Related Work}

\paragraph{State Space Models.}
SSMs \cite{gu2023mamba,mehta2022long,wang2023selective} are emerging architecture designs for sequence-to-sequence transformation. 
The design has the strength to model complex systems by focusing on how the input, output, and state variables evolve over time. Mamba-2~\cite{dao2024transformers} propose state space duality to design a new architecture whose core layer is a refinement of selective SSM. S4ND \cite{nguyen2022s4nd} is the first work that applies the state space mechanism to visual tasks and shows the potential to achieve  competitive performance with ViTs \cite{dosovitskiy2020image}. ViM~\cite{zhu2024vision} proposes a novel vision backbone with bidirectional selective SSM. The accomplishments 
demonstrate the potential of SSMs as an emerging foundation model family.  

\paragraph{Token Reduction.}
Token reduction is an effective strategy to enhance computational efficiency by reducing the number of processed tokens or patches \cite{modarressi2022adapler,huang2022pyramid,nawrot2022efficient,wang2023going,kong2023peeling,zhan2024exploringtokenpruningvision}. It enables significant acceleration without requiring additional weights or specialized hardware, aiming to selectively retain the most informative tokens. Several innovative approaches have been developed for Transformers. For example, EViT~\cite{liang2022evit} uses the attentiveness of the [CLS] token with respect to other tokens to identify the most important tokens. DynamicViT~\cite{rao2021dynamicvit} and SPViT~\cite{kong2021spvit} add layers that employ the Gumbel-Softmax trick to selectively prune less informative tokens.
Agile-Quant~\cite{shen2024agile} leverage the
activation-aware token pruning technique to reduce the outliers for LLMs.
ToMe \cite{bolya2022tome} measures dot product similarity between token keys to determine redundancy and merge accordingly. PuMer~\cite{cao2023pumer} proposed a token reduction framework for large-scale VLMs with text-informed pruning and modality-aware merging strategies to progressively reduce the tokens of input image and text. 

However, the dynamics of information flow between tokens and the learning mechanisms in models like Mamba \cite{gu2023mamba} remain largely unexplored. 
The absence of attention layers in Mamba makes current token reduction methods ineffective. Furthermore, 
the inclusion of the SSM module prevents the effective use of existing token reduction methods.

\section{Preliminary and Motivation}

\subsection{State Space Models}
SSMs are sequential models that map an input sequence \( x(t) \in \mathbb{R}^L \) to an output sequence \( y(t) \in \mathbb{R}^L \) through a hidden state \( h(t) \in \mathbb{R}^N \) as follows,
\begin{equation} 
    h'(t) = \mathbf{A} h(t) + \mathbf{B} x(t), \quad y(t) = \mathbf{C} h(t),
\label{eq:original_ssm}
\end{equation}
where \( L \) denotes the length of the sequence, \( N \) denotes the number of representation dimensions, \( \mathbf{A} \in \mathbb{R}^{N \times N} \) is the evolution matrix, and \( \mathbf{B} \in \mathbb{R}^{N \times L} \), \( \mathbf{C} \in \mathbb{R}^{L \times N} \) are the projection matrices.

Mamba ~\cite{gu2023mamba} represents a discrete version of the continuous system for SSMs and incorporates a timescale parameter \( \Delta \) to facilitate the transformation of continuous parameters with the zero-order hold (ZOH) as $\mathbf{\overline{A}} = \exp ( \Delta \mathbf{A} )$, and $\mathbf{\overline{B}} = (\Delta \mathbf{A})^{-1} (\exp(\Delta \mathbf{A}) - \mathbf{I}) \cdot \Delta \mathbf{B}.$
After obtaining the discretized \( \mathbf{\overline{A}} \) and \( \mathbf{\overline{B}} \), the discretization of Equation~(\ref{eq:original_ssm}) can be rewritten as,
\begin{equation} 
\begin{aligned}
\vspace{-2mm}
    \mathbf{h}_t = \mathbf{\overline{A}} \mathbf{h}_{t-1} + \mathbf{\overline{B}} \mathbf{x}_t, \quad \mathbf{y}_t = \mathbf{C} \mathbf{h}_t.
\end{aligned}
\label{eq:discret_mamba}
\end{equation}

Finally, the Mamba model computes the output through a global convolution as follows,
\begin{equation} 
\begin{aligned}
\vspace{-2mm}
    \mathbf{\overline{K}} &= ( \mathbf{C} \mathbf{\overline{B}}, \mathbf{C} \mathbf{\overline{AB}}, \ldots, \mathbf{C} \mathbf{\overline{A}^{L-1}\overline{B}} ), \\
    \mathbf{y} &= \mathbf{x} * \mathbf{\overline{K}},
\end{aligned}
\label{eq:mamba_kernel}
\end{equation}
where \( \mathbf{y} \) denotes the output sequence, \( L \) denotes the length of the input sequence \( \mathbf{x} \), and \( \mathbf{\overline{K}} \in \mathbb{R}^L \) denotes a structured convolutional kernel.

\begin{figure}[t]
    \centering
    \includegraphics[width=0.99\columnwidth]{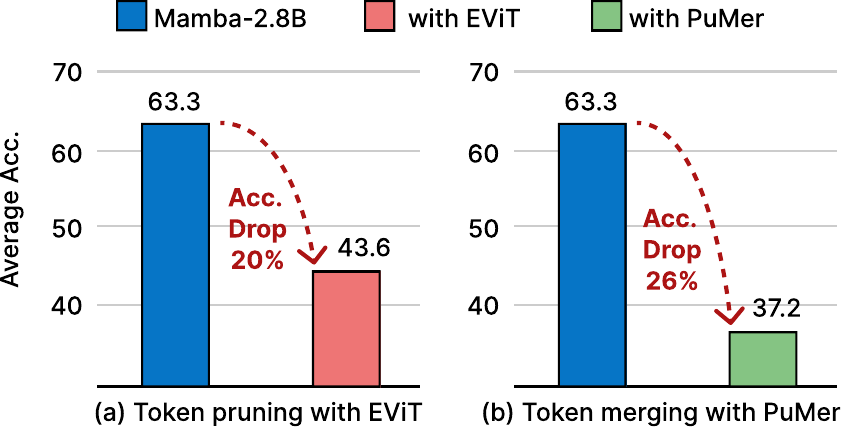}
    \caption{Performance of applying token pruning (EViT) and merging (PuMer) methods on Mamba-2.8B, showcasing significant drop in accuracy.}
    \label{fig:previous}
    \vspace{-4mm}
\end{figure}

\subsection{Analysis of Reasons Behind the Failure of Token Reduction on SSMs} \label{sec:fail}
Due to the SSMs' reliance on a sequential strategy for token computation, the previous token reduction strategies highlighted in Figure~\ref{fig:previous} do not yield effective results. In this section, we  delve into the reasons why directly applying  SOTA token pruning or merging method fails on SSMs. 

\paragraph{Failure of token pruning on SSMs.} Existing SOTA token pruning methods for Transformers, such as Token Filtering~\cite{berchansky2023optimizing}, Agile-Quant~\cite{shen2024agile}, and EViT~\cite{liang2022evit}, typically involve sorting all tokens in the current layer based on an 
importance evaluation criterion, and then removing the less important tokens. 
As shown in Figure~\ref{fig:previous}(a), after we directly implement post-training token pruning (EViT) to reduce 20\% of the overall FLOPS for   Mamba-2.8B,  there is a dramatic drop in average accuracy on zero-shot evaluation. This performance drop is introduced by pruning certain tokens with \textit{unrecoverable information loss}, although the pruned tokens are less important based on a heuristic importance  metric. This \textit{information loss} is \textit{gradually amplified} during the sequence computations process of Equation~(\ref{eq:discret_mamba}) and (\ref{eq:mamba_kernel}) in SSMs. 


\paragraph{Failure of token merging on SSMs.} On the other hand, linguistic contexts often contain \textit{redundant} tokens, which 
do not add significant contextual depth to the model's understanding. ToMe~\cite{bolya2022tome} introduces a bipartite token merging strategy for vision Transformers. Following this, initiatives like PuMer~\cite{cao2023pumer} extend this strategy to vision-language models, merging redundant tokens in linguistic model components and their vision counterparts at the same time. However, as shown in Figure~\ref{fig:previous}(b), applying this bipartite token merging strategy directly to SSMs proves ineffective. The strategy uniformly partitions the tokens in the current layer into two groups, and merges tokens in one group into the other group,  
disregarding the inherent value (or token importance) of each token.  Thus, certain important tokens may be merged into other tokens.  Given the critical role of important tokens in sequence computations using Equation~(\ref{eq:mamba_kernel}) in SSMs, \textit{overlooking} the \textit{inherent significance} of tokens and thus removing important tokens can lead to 
substantially different $\mathbf{y}$ in Equation~(\ref{eq:mamba_kernel}) and thus severe performance degradation.



\subsection{Motivation}
From the analysis presented, we conclude that the failure of token pruning in SSMs comes from the loss of crucial information due to token removal. Meanwhile, the failure of token merging in SSMs can be attributed to the neglect of token importance. This oversight can result in a more significant drop in accuracy compared to pruning, underscoring the critical role of token importance in the model’s performance. 
Therefore, our objective is to combine token importance and similarity as guidance for a unified token reduction method (combining pruning and merging). We aim to develop a more fine-grained reduction strategy to handle the computation sensitivity of selective SSMs, ensuring that the reduction process maintains model accuracy and efficiency simultaneously. 

\section{Methodology}

To tackle the problem, we first rethink the token importance  metric for SSMs. We then introduce a novel approach for unified token reduction by token importance classification  that combines the advantages of both token pruning and token merging to facilitate faster and memory-efficient computation across SSM layers. 

\subsection{Rethinking Token Importance Metric for State Space Models}

\label{sec:pruning_metric}
To derive the appropriate token importance metric, we look at the layer computations in SSMs such as Mamba. For the $l^{th}$ layer, the input token sequence $\mathbf T_{l-1} \in \mathbb{R}^{B \times N \times D}$ is first projected to $\mathbf{x}\in \mathbb{R}^{B \times N \times D'}$, and then goes through SSMs for data-dependent context modeling. It processes  $\mathbf{x}$ from the \texttt{forward} scan via:
\begin{equation}
\begin{aligned}
        \mathbf{y} \leftarrow & \texttt{SSM}(\mathbf A, \mathbf B, \mathbf C)(\mathbf{x}), \\ 
\end{aligned}
\end{equation}
where the hidden states $\mathbf{y} \in \mathbb{R}^{B \times N \times D'}$ is the output of  SSM  (see Equation \eqref{eq:mamba_kernel}). 
The token sequence  output of the  $l^{th}$ layer can be obtained as $\mathbf T_l \leftarrow \texttt{Linear}^T \mathbf{y} + \mathbf T_{l-1}.$
To evaluate the importance of each token, we first extract the hidden states $\mathbf{y}$ from the SSM layer, denoted as $\mathbf{y}\in \mathbb{R}^{B \times N \times D'}$. The hidden states represent the intermediate representations of the tokens after passing through the SSM layer.
To quantify the importance of each token, we compute the sum of the $\mathbf{y}$ across the last dimension, which corresponds to the feature dimension $D'$. 
The SSMs architecture, with its high-dimensional channel space, allows for a finer-granularity analysis of attention across numerous channels. Unlike Transformers that produce a single attention matrix per head, SSMs exploit their extensive channel capacity for a more detailed attention distribution, enhancing the model's ability to discern subtle features and interactions among tokens. Thus, we aggregate the clipped values across all channels for each token to evaluate token importance as follows,
\begin{equation} \label{eq:importance_metric}
    \mathcal{S} = \frac{\sum_{d=1}^{D'} \texttt{max}(0,[\mathbf{y}]_{::d})}{D'},
\end{equation}
where $[\cdot]_{::d}$ denotes the $d^{th}$ feature map in the  feature dimension with size $D'$.
We use $\mathcal{S} \in \mathbb{R}^{B \times N \times 1}$ as the token importance metric corresponding to $B\times N$ tokens to guide the reduction process, ensuring that only the most contextually relevant tokens are retained. 
To make a comprehensive study, we compare the performance with other token importance metrics, including the $\ell_{1}$ norm, $\ell_{2}$ norm, as well as unclipped values without the \texttt{max} operation. We find that using clipped values in Equation  \eqref{eq:importance_metric} as the token importance metric can constantly yield better results.

\begin{figure*}[t!]
    \centering
    \includegraphics[width=0.99\linewidth]{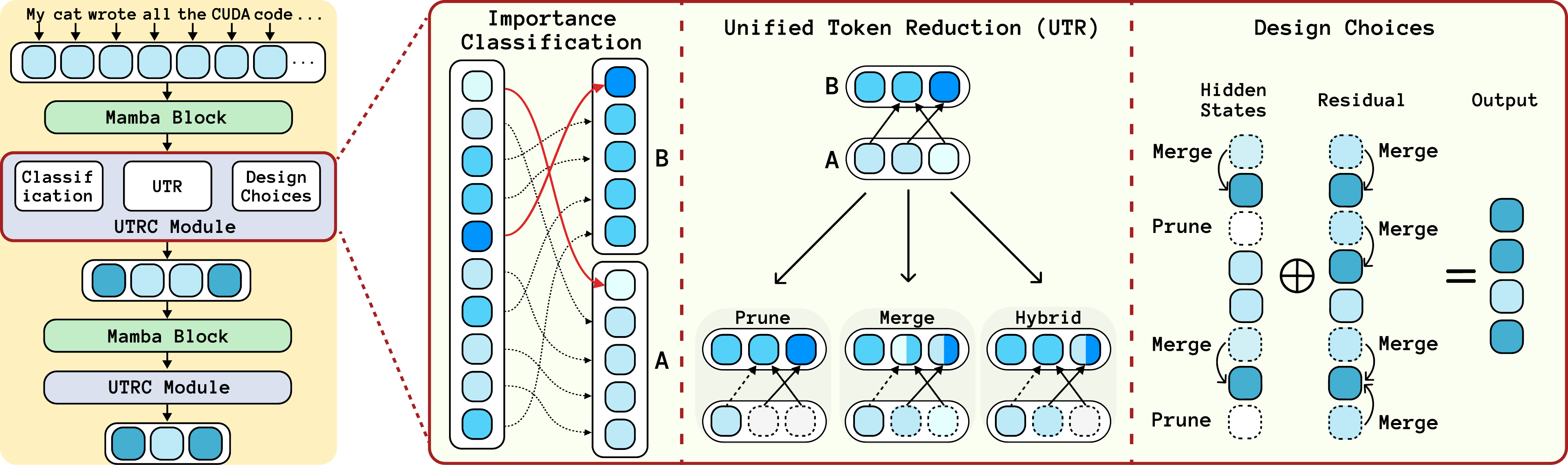}
    \caption{Overview of our proposed Unified Token Reduction by token importance Classification (UTRC) method. It contains three parts: Token Importance Classification, Unified Token Reduction (UTR), and Design Choices. Lighter colors indicate tokens with less importance, and darker colors indicate tokens with greater importance.
    }
    \label{fig:overview}
    \vspace{-3mm}
\end{figure*}
\subsection{Unified Token Reduction by Token Importance Classification}

To achieve token reduction, it is important to derive a token importance classification strategy that effectively differentiates between less important and more important tokens.
However, it is challenging to directly classify thousands of tokens in real-time due to high complexity. 
To overcome this, we further leverage the token importance evaluation as in Equation \eqref{eq:importance_metric}, and employ a decoupling strategy. The strategy initially computes the importance of each token, followed by classification based on this obtained importance. After that, we perform unified token reduction (UTR) and leverage multiple design choices to enable effective and fine-grained strategies. Figure~\ref{fig:overview} illustrates our proposed approach. 
The steps of our method are as follows:

\begin{enumerate}[leftmargin=*, itemsep=-2pt,topsep=5pt]
    \item \textbf{Calculate}   token importance with Equation \eqref{eq:importance_metric}.
    \item \textbf{Classify} the tokens into set $M_A$ and $M_B$ based on their importance. At the end, $N/2$ less important tokens are assigned to set $M_A$, with the rest $N/2$ more important tokens to set $M_B$. 
    \item \textbf{Create} a single connection from each token in set $M_A$ to its most similar counterpart in set $M_B$, as shown below, 
    \begin{equation}
          f_{i} = \underset{b_j \in M_B} {\arg\max}
          \ \text{sim}(a_i, b_j),
    \end{equation}
    \begin{equation}
          g_{i} = \underset{b_j \in M_B} {\max}
          \ \text{sim}(a_i, b_j),
    \end{equation}
    where  $\text{sim}(a, b)$ is the cosine similarity between token $a$  and $b$, $f_{i}$ denotes the most similar token in  $M_B$ to $a_i \in M_A $, and $g_i$ is the corresponding largest similarity between $a_i $ and   $ f_{i}$.   
    \item \textbf{Retain} the $p\%$ most similar connections after sorting $\{g_i, \forall i\}$. 
    \item \textbf{Process} the connected tokens with our UTR method.
    \item \textbf{Reassemble} the two sets of tokens into one set.
\end{enumerate}


\paragraph{Unified token merging and pruning.} 
For the $5^{th}$ step of our method, we apply two token reduction strategies -- merging and pruning. We can apply token pruning or merging for each of the connections obtained from the $4^{th}$ step. For token pruning, we do not change the tokens in Set $M_B$ and 
only update  Set $M_A$ by removing the token $a_i$, i.e.,  $M_A = M_A\setminus a_i$, where $\setminus$ denotes the operation of element removal from the set. Consequently, $f_i$ represents the remaining token in $M_B$ for a connected pair $(a_i, f_i)$. For merging, the tokens connected by retained pairs are combined by averaging their features. Specifically, we update the most similar token in $M_B$ with $f_i = (a_i + f_i) / 2$, and  remove $a_{i}$ from $M_A$. The modified  ${f_i}$   represents the fused token for the connected pair $(a_i, f_i)$. 

Our proposed merging and pruning techniques can be seamlessly integrated as shown in the UTR part in Figure \ref{fig:overview}. This allows for fine-grained reduction strategies across intra-layer branches, enabling distinct reduction strategies to both hidden states and residuals. The motivation is to address the \textit{removed index misalignment} issue between branches. Such misalignment occurs when a token reduced in the hidden state is not concurrently reduced in the residual branch, and vice versa. This discrepancy, especially when branches recombine at the end of each layer, can significantly lower the overall compression ratio and hinder the effectiveness of fine-grained token reduction strategies. By unifying these techniques, we can optimize the method while meeting the required compression levels.



\paragraph{Hybrid token reduction.} 

With the proposed UTR strategy, we further leverage a fine-grained strategy to balance the information importance and redundancy.
For the corresponding tokens of retained $p\%$ most similar connections (the $4^{th}$ step), we prune $(p \times q)\%$ tokens and merge the remaining $[p\times(1-q)]\%$ tokens. We find that $q=0.5$ leads to best performance compared with other $q$ values. We provide a detailed evaluation in Table \ref{table:choices}.

\subsection{Design Choices}

\paragraph{Intra-layer token reduction design.} We delve deeper into our intra-layer token reduction design tailored for SSMs, targeting the hidden states and residual connections. Our approach employs the \textit{hybrid token reduction} strategy on \textit{hidden state} tokens, meticulously designed to strike a balance between preserving essential information and eliminating redundancy. By discerning the contextual significance of each token, this strategy focuses on removing tokens with minimal contextual relevance, thus enhancing the overall informational flow of the SSM module. This design choice not only preserves but also amplifies the high-contextual tokens.
\textit{Residual connections} are crucial for maintaining the integrity of information from the last layer. Therefore, we aim to preserve as much residual information as possible through our \textit{token merging} method. The final design is shown in the design choices part in Figure \ref{fig:overview}.
Empirical results support our fine-grained design, demonstrating that reducing tokens with our method in the hidden state and residual connection areas effectively preserves the performance of SSMs.

\begin{table*}[t!]
\centering
\resizebox{0.99\linewidth}{!}{
\begin{tabular}{l|c|c|cccccc|c}
\toprule
\multirow{2}{*}{Method} & FLOPS         & \multicolumn{2}{c}{LAMBADA} & HellaSwag & PIQA & Arc-E & Arc-C & WinoGrade & Avg. \\
\cline{3-10}
                        & Reduction      & PPL $\downarrow$    & Acc$\uparrow$(\%)    & Acc$\uparrow$(\%)       & Acc$\uparrow$(\%)  & Acc$\uparrow$(\%)   & Acc$\uparrow$(\%)   & Acc$\uparrow$(\%)       & Acc$\uparrow$(\%)     \\
\toprule
Mamba-2-1.3B              & 0\% &  5.02 & 65.7 & 59.9 &  73.2&  64.3 & 33.3 & 60.9 &  59.5    \\

\midrule
+ PuMer           & \multirow{3}{*}{10\%}             &532.52 &33.3 &27.5 &61.3 &57.8 &30.6 &59.8 &45.1     \\
+ EViT               &              &27.10 &52.2 &32.9 &68.9 &63.2 &33.2 &61.0 &51.9         \\
\rowcolor{blue!12} + \textbf{Ours}               &             &11.16 &55.9 &59.2 &71.0 &64.3 &34.1 &61.0 &57.6        \\
\midrule
+ PuMer           & \multirow{3}{*}{20\%}             &49017.23 &14.9 &25.5 &54.1 &45.5 &28.2 &54.4 &37.1      \\
+ EViT               &              &1655.76 &32.4 &26.5 &59.4 &56.9 &30.6 &59.4 &44.2         \\
\rowcolor{blue!12} + \textbf{Ours}               &             &25.94 &46.1 &58.0 &64.3 &64.0 &34.4 &60.7 &54.6        \\
\toprule
Mamba-2-2.7B              & 0\% & 4.10  & 69.7  & 66.6   & 76.4   & 69.6   & 36.4   & 64.0   & 63.8     \\
\midrule
+ PuMer           & \multirow{3}{*}{10\%}             &712.73 &36.4 &27.2 &63.4 &63.8 &30.9 &63.5 &47.5     \\
+ EViT               &             &11.43 &55.8 &35.7 &72.0 &69.1 &35.4 &64.1 &55.4        \\
\rowcolor{blue!12} + \textbf{Ours}               &            &8.55 &59.0 &66.1 &73.2 &69.4 &36.5 &64.0 &61.4        \\
\midrule
+ PuMer           & \multirow{3}{*}{20\%}             &7820.51 &20.7 &25.9 &56.0 &50.5 &28.8 &56.0 &39.7      \\
+ EViT               &             &196.42 &44.5 &28.8 &65.1 &62.3 &32.6 &63.9 &49.6         \\
\rowcolor{blue!12} + \textbf{Ours}               &             &17.96 &49.1 &64.7 &68.2 &69.4 &37.5 &63.1 &58.7       \\
\midrule
+ PuMer           & \multirow{3}{*}{30\%}             &49301.49 &10.6 &26.9 &53.9 &44.4 &29.2 &53.5 &36.4      \\
+ EViT               &             &3412.13 &27.9 &25.9 &57.7 &51.8 &27.3 &59.1 &41.6         \\
\rowcolor{blue!12} + \textbf{Ours}               &            &42.61 &38.3 &59.4 &61.2 &68.4 &37.3 &63.9 &54.7       \\
\bottomrule
\end{tabular}
}
\caption{Main results of post-training performance on Mamba-2-1.3B and Mamba-2-2.7B. We compare with baseline methods and evaluate them on six benchmarks under 10\%, 20\%, and 30\% FLOPS reduction.}
\label{tab:main_results2}
\vspace{-0.4cm}
\end{table*}

\paragraph{Hierarchical token reduction procedure.} 

We apply a hierarchical method to reduce tokens across multiple layers. Tokens reduced in one layer are further reduced in subsequent layers, balancing overall efficiency and performance. Reducing tokens in each layer can cause high overhead, as token importance between adjacent layers is often similar. Thus, it is unnecessary to reduce tokens at every layer. 
Furthermore, reducing tokens in earlier layers yields greater computational savings, but these layers cannot fully capture token importance. In our experiments, we apply token reduction after at least the $10^{th}$ layer and every 5 layers with a fixed compression ratio.

\section{Experiment Results}

\subsection{Implementation Details} \label{sec:setting}
We implement our method based on   PyTorch \cite{paszke2019pytorch}     for scientific computations  and   HuggingFace \cite{wolf2019huggingface}   for managing models. We use  Mamba models to test the effectiveness of our method. Our approach covers a variety of Mamba models, with Mamba-2-2.7B, Mamba-2-1.3B, Mamba-2.8B and Mamba-1.4B. 
We evaluate the task performance on multiple common sense reasoning datasets including LAMBADA~\cite{paperno2016lambada}, HellaSwag~\cite{zellers2019hellaswag}, PIQA~\cite{bisk2020piqa}, Arc-easy~\cite{clark2018arc}, Arc-challenge~\cite{clark2018arc}, and WinoGrade~\cite{sakaguchi2021winogrande}.
Perplexity on LAMBADA dataset and average accuracy on all mentioned datasets are provided. All experiments are conducted on a NVIDIA A100 80GB GPU. 

\paragraph{Reduction locations.} We adopt the hierarchical token reduction procedure. For Mamba2-2.7B and Mamba-2.8B, we perform all methods in the [12, 17, 22, 27, 32, 37, 42] layers; for Mamba2-1.3B and Mamba-1.4B, we perform all methods in the [10, 15, 20, 25, 30, 35] layers.  We use a fixed compression ratio for each prune layer.

\paragraph{Evaluation Details.}
The evaluation of perplexity (PPL) and average accuracy are adjusted to account for the reduction in the number of output due to token reduction. The target label logits are adjusted accordingly. For example, when the output token reduction rate is $m\%$, the label logits are also reduced to their first $1-m\%$ logits to calculate the PPL and average accuracy properly.

\paragraph{Baselines.} We compare our method with PuMer \cite{cao2023pumer} and EViT \cite{liang2022evit}. PuMer, which includes a dedicated text token reduction module, can be directly adopted in our study. For EViT, originally designed for vision Transformers, we configure it to ensure a fair comparison in our evaluation.

\begin{table*}[t!]
\centering
\resizebox{0.99\linewidth}{!}{
\begin{tabular}{l|c|c|cccccc|c}
\toprule
\multirow{2}{*}{Method} & FLOPS         & \multicolumn{2}{c}{LAMBADA} & HellaSwag & PIQA & Arc-E & Arc-C & WinoGrade & Avg. \\
\cline{3-10}
                        & Reduction      & PPL $\downarrow$    & Acc$\uparrow$(\%)    & Acc$\uparrow$(\%)       & Acc$\uparrow$(\%)  & Acc$\uparrow$(\%)   & Acc$\uparrow$(\%)   & Acc$\uparrow$(\%)       & Acc$\uparrow$(\%)     \\
\toprule
Mamba-1.4B              &  0\% & 5.04   & 64.9   & 59.1      & 74.2 & 65.5  & 32.8  & 61.5      & 59.7    \\
\midrule
+ PuMer           & \multirow{3}{*}{10\%}             &534.91 &34.6 &25.8 &59.7 &55.6 &29.5 &59.5 &44.1     \\
+ EViT               &              &43.69 &47.6 &33.0 &69.2 &64.3 &32.1 &61.4 &51.3         \\
\rowcolor{blue!12} + \textbf{Ours}               &            &11.46 &56.5 &58.9 &71.3 &65.1 &33.9 &61.4 &57.8        \\
\midrule
+ PuMer           & \multirow{3}{*}{20\%}             &11733.02 &13.1 &25.6 &52.5 &41.8 &27.2 &48.8 &34.8      \\
+ EViT               &              &5687.80 &21.8 &26.3 &58.4 &54.0 &28.2 &58.2 &41.1         \\
\rowcolor{blue!12} + \textbf{Ours}               &            &31.32 &44.9 &57.7 &62.8 &62.8 &33.2 &59.0 &53.4        \\
\toprule
Mamba-2.8B              & 0\% & 4.23   & 69.2   & 66.1      & 75.2 & 69.7  & 36.3  & 63.5      & 63.3    \\
\midrule
+ PuMer           & \multirow{3}{*}{10\%}            &487.09 &36.6 &26.3 &62.4 &63.6 &30.7 &63.1 &47.1      \\
+ EViT               &              &174.92 &51.8 &35.7 &71.0 &68.9 &35.7 &63.2 &54.4         \\
\rowcolor{blue!12} + \textbf{Ours}               &            &9.53 &59.9 &66.0 &72.0 &69.8 &36.7 &63.5 &61.3       \\
\midrule
+ PuMer           & \multirow{3}{*}{20\%}             &10746.15 &17.9 &25.3 &52.5 &47.0 &28.7 &52.0 &37.2      \\
+ EViT               &             &9784.73 &26.9 &24.8 &59.9 &57.2 &29.9 &63.1 &43.6         \\
\rowcolor{blue!12} + \textbf{Ours}               &             &23.97 &49.0 &63.8 &62.3 &68.5 &38.1 &64.0 &57.6        \\

\midrule

+ PuMer           & \multirow{3}{*}{30\%}             &140763.76 &6.0 &26.0 &54.6 &41.5 &26.6 &51.7 &34.4     \\
+ EViT               &              &63230.76 &12.3 &25.0 &52.5 &41.9 &23.6 &51.9 &34.5         \\
\rowcolor{blue!12} + \textbf{Ours}               &             &81.16 &36.1 &39.4 &58.1 &66.2 &37.1 &60.8 &49.6       \\

\bottomrule
\end{tabular}
}
\caption{Main results of post-training performance on Mamba-1.4B and Mamba-2.8B. We compare with baseline methods and evaluate them on six benchmarks under 10\%, 20\%, and 30\% FLOPS reduction.}
\label{tab:main_results}
\vspace{-0.4cm}
\end{table*}

\subsection{Quantitative Evaluation}

\paragraph{Evaluation on Mamba-2.}
As shown in Table~\ref{tab:main_results2}, for Mamba-2 models (1.3B and 2.7B), our method consistently achieves better performance than all baselines (PuMer and EViT) with non-marginal improvements under the same FLOPS reduction ratios.  For Mamba-2-1.3B,  our method achieves significantly lower PPL and higher accuracy on almost all downstream datasets, with an average accuracy 10\% (54.6\% v.s. 44.2\% from EViT) higher than the best baseline under 20\% FLOPS reduction. For Mamba-2-2.7B, our method outperforms baselines on various benchmarks with wide margins, achieving an average accuracy  13.1\% higher than the best baseline under 30\% FLOPS reduction. 


\paragraph{Evaluation on Mamba.}
As demonstrated in Table~\ref{tab:main_results}, for Mamba models (1.4B and 2.8B), we can make similar observations that our method outperforms all baselines with non-marginal improvements in terms of PPL and accuracy on multiple benchmarks. Our method maintains a low PPL while baselines can hardly keep a reasonable PPL (such as our 23.97 PPL v.s. 9785 from EViT under 20\% FLOPS reduction for Mamba-2.8B). Our average accuracy is significantly higher than baselines, such as our 53.4\%  over 41.1\% from EViT for Mamba-1.4B  under 20\% FLOPS reduction.  


\paragraph{Summary.}
For SSMs such as Mamba, our proposed method consistently demonstrates better performance in terms of PPL and average accuracy across various levels of FLOPS reduction compared with baselines.  PuMer and EViT fail to maintain high performance due to the reasons discussed in Section~\ref{sec:fail}. After an insightful investigation of the reasons for failure and a comprehensive design to combine the advantages of pruning and merging, our unified method can effectively and efficiently prune tokens in SSMs without significant performance degradation. 





\subsection{Ablation Study \& Analysis}

\paragraph{Different Importance Metric.}
We study the token importance metric for our token reduction strategy. As shown in Table~\ref{tab:metrics}, for 
Mamba-2-2.7B and Mamba-2.8B, we provide a comparative analysis of different metrics: $\ell_1$-norm, $\ell_2$-norm, without \texttt{Clip} (the max function in Equation \eqref{eq:importance_metric}), and with \texttt{Clip}, along with their impacts on LAMBADA PPL and average accuracy across six tasks (as in Table~\ref{tab:main_results}). The results show that \texttt{Clip} achieves the lowest PPL of 17.96 and the highest average accuracy of 58.7\% for Mamba-2-2.7B, outperforming other metrics. For  Mamba-2.8B, though \texttt{Clip} has a slightly higher PPL, its average accuracy is the highest 57.6\%. 
This analysis underscores the importance of the proposed token importance metric in enhancing model accuracy and efficiency.

\begin{table}[t]
\centering
\resizebox{0.99\linewidth}{!}{\begin{tabular}{l|l|c|c}
\toprule
Model & Metric & \begin{tabular}{@{}c@{}}LAMBADA \\ PPL $\downarrow$ \end{tabular} & \begin{tabular}{@{}c@{}}Avg. \\ Acc.  $\uparrow$(\%) \end{tabular} \\ \midrule
\multirow{4}{*}{Mamba-2-2.7B}& $\ell_{1}$-norm  & 17.96 & 58.6   \\
&$\ell_{2}$-norm &  19.86 & 58.6  \\
&w/o Clip  & 18.17 & 58.5   \\
&\cellcolor{blue!12}\bf Clip (ours) &\cellcolor{blue!12}\bf  17.96  & \cellcolor{blue!12}\bf 58.7   \\  \midrule
\multirow{4}{*}{Mamba-2.8B}& $\ell_{1}$-norm  & \bf 23.93 & 56.8   \\
&$\ell_{2}$-norm & \bf 23.93  & 57.5   \\
&w/o Clip  & 1365.69 & 40.7   \\
&\cellcolor{blue!12}\bf Clip (ours) & \cellcolor{blue!12}23.97 & \cellcolor{blue!12}\bf 57.6   \\ 
\bottomrule
\end{tabular}}
\caption{Ablation study of token importance metric with our unified token merging and pruning design.}
\label{tab:metrics}
\vspace{-0.4cm}
\end{table}

\paragraph{Reduction location analysis.}
The choice of token reduction location impacts model performance.
Table~\ref{tab:location} presents the ablation study of reduction location on  Mamba-2-2.7B under a 20\% FLOPS reduction. Notably, the configuration with reduction layers at [12, 17, 22, 27, 32, 37, 42] achieves the lowest PPL 17.96 and the highest 58.7\% average accuracy, demonstrating the effectiveness of this specific reduction strategy. In contrast, deeper reduction layers, such as [20, 25, 30, 35, 40, 45, 50], result in higher PPL and lower average accuracy, indicating that deeper layers do not always yield better results. 
Token reduction at earlier layers can lead to higher computation  efficiency 
without sacrificing accuracy significantly.

\begin{table}[t]
\centering
\resizebox{0.95\linewidth}{!}{\begin{tabular}{c|cc}
\toprule
\begin{tabular}{@{}c@{}}Location \\ (every 5 layers) \end{tabular} & \begin{tabular}{@{}c@{}}LAMBADA \\ PPL $\downarrow$ \end{tabular} & \begin{tabular}{@{}c@{}}Avg. \\ Acc. $\uparrow$(\%) \end{tabular} \\ \midrule
$[20,25,30,35,40,45,50]$ & 18.88  & 57.8   \\
$[18,23,28,33,38,43,48]$ & 18.32   & 58.3   \\ 
$[16,21,26,31,36,41,46]$ & 18.79  & 58.1   \\ 
$[14,19,24,29,34,39,44]$ & 18.74  & 58.3   \\  
$[10,15,20,25,30,35,40]$ & 18.76    & 58.2   \\ 
\rowcolor{blue!12} $[12,17,22,27,32,37,42]$ & \bf 17.96  & \bf 58.7   \\  
\bottomrule
\end{tabular}}
\caption{Ablation study of reduction location on Mamba-2-2.7B under 20\% overall reduction of FLOPS.}
\label{tab:location}
\vspace{-0.2cm}
\end{table}

\paragraph{Different design choices.} 
For hidden states and residual connections, we can apply pruning, merging, or our hybrid token reduction with different combinations of pruning and merging (denoted by $q$). We conduct ablation studies to find the optimal $q$ configuration for both hidden states and residual connections. 
Table~\ref{table:choices} presents experiments on the Mamba-2-2.7B model under a 30\% FLOPS reduction. The results indicate that the combination of $q=0.5$ for hidden states and merging only for residual connections achieves the lowest 40.61 PPL   and the highest 54.7\% average accuracy, highlighting its effectiveness in this context. 
Furthermore, combining pruning and merging with $q=0.5$  for hidden states consistently outperforms pruning-only or merging-only strategies. Notably, even our basic method using importance classification (M-only \& M-only Acc. 54.0\%) outperforms existing methods (PuMer Acc. 36.4\% and EViT Acc. 41.6\%) by a large margin.


\begin{figure}[t]
    \centering
    \includegraphics[width=0.74\columnwidth]{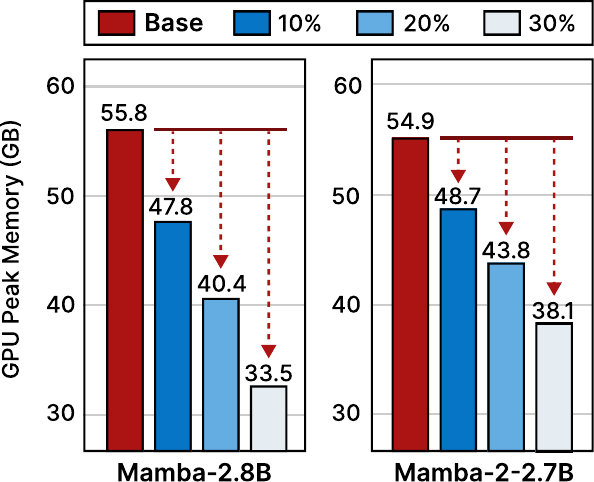}
    \caption{Comparison of GPU peak memory reduction between different FLOPS reduction ratios for Mamba-2.8B and Mamba-2-2.7B.}
    \label{fig:memory}
    \vspace{-4mm}
\end{figure}

\subsection{Efficiency Results}
\label{sec:efficient_results}
\begin{table}[t]
\centering 
        \resizebox{0.93\linewidth}{!}{
        \begin{tabular}{cc|ccc}
        \toprule 
         \begin{tabular}{@{}c@{}}Hidden \\ States \end{tabular} & \begin{tabular}{@{}c@{}}Residual \\ Connections \end{tabular}  &  \begin{tabular}{@{}c@{}}LAMBADA \\ PPL $\downarrow$ \end{tabular} & \begin{tabular}{@{}c@{}}Avg. \\ Acc. $\uparrow$(\%) \end{tabular}    \\
        \midrule
        M-only & M-only &42.61&  54.0 \\
        P-only & P-only &  42.65 &  53.9 \\
        $q=0.8$ & $q=0.2$ & 42.65 &  54.3 \\
        $q=0.2$ & $q=0.8$ & 42.67 &  54.1 \\
        $q=0.5$ & $q=0.5$ &  42.35 & 53.7 \\
        $q=0.5$ & P-only &  42.67  &  54.1 \\
        \rowcolor{blue!12} $q=0.5$ &  M-only & \bf 40.61  & \bf 54.7 \\
        \bottomrule
        \end{tabular}
        }
        \captionof{table}{Ablation study of different design choices on Mamba-2-2.7B under 30\% overall reduction of FLOPS.}
         \label{table:choices}
         \vspace{-0.2cm}
\end{table}

\begin{figure}[t]
    \centering
    \includegraphics[width=0.97\columnwidth]{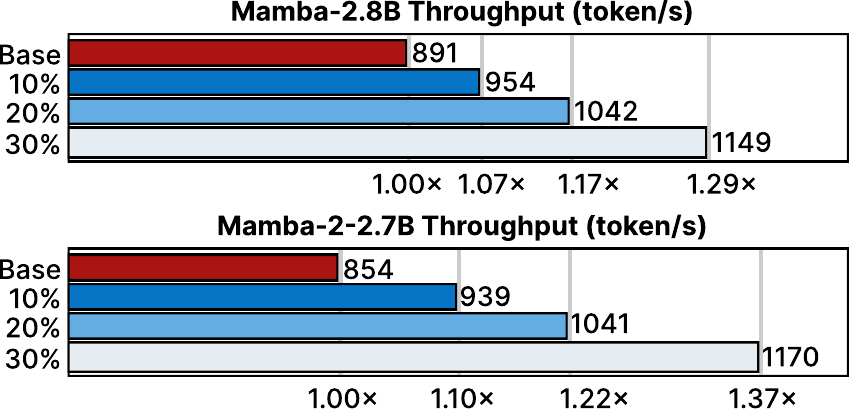}
    \caption{Comparison of the generation throughput between different FLOPS reduction ratios for Mamba-2.8B and Mamba-2-2.7B.}
    \label{fig:throughput}
    \vspace{-4mm}
\end{figure}

We evaluate the GPU peak memory usage of Mamba-2.8B and Mamba-2-2.7B when generating 2048 tokens with a batch size 96 under various FLOPS reduction ratios.
As illustrated in Figure \ref{fig:memory},
the GPU peak memory reduction for Mamba-2.8B can reach up-to 14.4\%, 27.7\%, and 40.0\%, under 10\%, 20\%, and 30\% FLOPS reduction, respectively. For Mamba-2-2.7B, it can reduce the peak memory by 11.4\%, 20.3\%, 30.6\% when reducing 10\%, 20\%, and 30\% FLOPS, respectively.

Further, our proposed method can lead to practical inference acceleration with higher model throughput, as shown in Figure \ref{fig:throughput}.
The throughput can be improved by $1.07\times$, $1.17\times$, and $1.29\times$ for Mamba-2.8B, and $1.10\times$, $1.22\times$, and $1.37\times$ for Mamba-2-2.7B, when reducing 10\%, 20\%, and 30\% FLOPS, respectively.
The throughput measurements are collected with a batch size 16 by generating 100 tokens with a prompt length of 2048. 
More details and efficiency results of other models can be found in Appendix \ref{sec:appendix}.

\section{Conclusion}

In this paper, we introduced a unified post-training token reduction method for SSM architectures like Mamba. We addressed the limitations of existing token reduction techniques by combining token importance and similarity to create a fine-grained reduction strategy. 
Our method includes multiple design choices for effective intra-layer optimizations. 
Experiments show significant reductions in computational demands and peak memory usage, while maintaining competitive accuracy, outperforming baseline methods on benchmarks.

\section*{Limitations}
Our experiments do not involve results after fine-tuning, which we believe could further improve the performance of our method. While our approach is applicable to Transformer-based LLMs, we have not tested it on other Transformer-based LLMs. We intend to address these extensions in future work.

\section*{Acknowledgement}
This work is supported by National Science Foundation CNS-2312158. We would like to express our sincere gratitude to the reviewers for their invaluable feedback and constructive comments to improve the paper.

\bibliography{acl_latex}

\newpage
\appendix

\section{Appendix}
\label{sec:appendix}

\subsection{More Details}
Peak memory refers to the maximum memory required during a program's execution. If the peak memory exceeds the available VRAM on a GPU, it will result in an \textbf{``Out of Memory''} error, preventing the program from running.

\subsection{More Efficiency Results}

\begin{figure}[h]
    \centering
    \includegraphics[width=0.75\columnwidth]{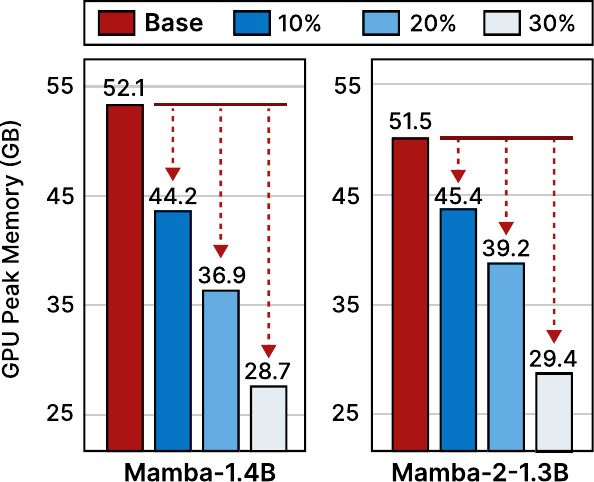}
    \caption{Comparison of GPU peak memory reduction between different FLOPS reduction ratios for Mamba-1.4B and Mamba-2-1.3B.}
    \label{fig:memory-small}
\end{figure}

The GPU peak memory usage of Mamba-1.4B and Mamba-2-1.3B are shown in Figure \ref{fig:memory-small} following the same configuration as Section \ref{sec:efficient_results}.
We follow the PyTorch instruction\footnote{\url{https://pytorch.org/docs/stable/torch_cuda_memory.html}} to capture the GPU peak memory snapshot.

When reducing 10\%, 20\%, and 30\% FLOPS compared to the baseline, Mamba-1.4B can obtain up to 15.2\%, 29.1\%, and 44.7\% peak memory reduction, while the peak memory reduction for Mamba-2-1.3B can reach up-to 11.9\%, 23.9\%, and 42.9\%.

\begin{figure}[h]
    \centering
    \includegraphics[width=0.99\columnwidth]{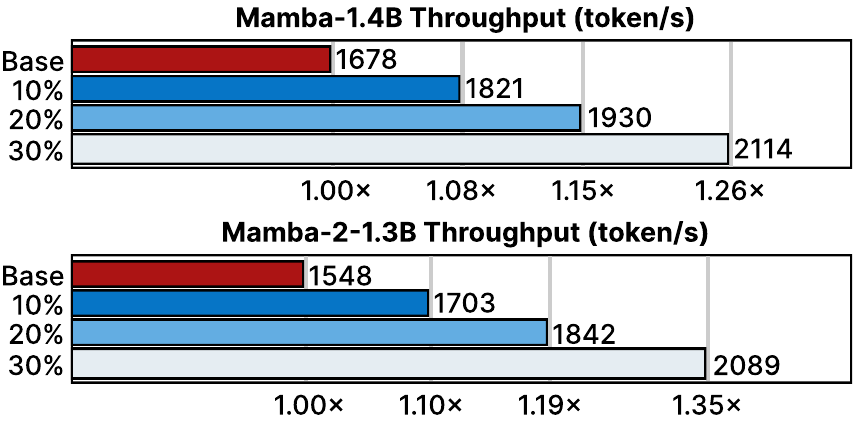}
    \caption{Comparison of the generation throughput between different FLOPS reduction ratios for Mamba-1.4B and Mamba-2-1.3B.}
    \label{fig:throughput-small}
\end{figure}
\begin{table*}[h!]
\centering
\resizebox{0.99\linewidth}{!}{
\begin{tabular}{l|c|c|cccccc|c}
\toprule
\multirow{2}{*}{Method} & FLOPS         & \multicolumn{2}{c}{LAMBADA} & HellaSwag & PIQA & Arc-E & Arc-C & WinoGrade & Avg. \\
\cline{3-10}
                        & Reduction      & PPL $\downarrow$    & Acc$\uparrow$(\%)    & Acc$\uparrow$(\%)       & Acc$\uparrow$(\%)  & Acc$\uparrow$(\%)   & Acc$\uparrow$(\%)   & Acc$\uparrow$(\%)       & Acc$\uparrow$(\%)     \\
\toprule
Mamba-2-2.7B              & 0\% & 4.10  & 69.7  & 66.6   & 76.4   & 69.6   & 36.4   & 64.0   & 63.8     \\
\midrule
+ LTMP           & \multirow{2}{*}{10\%}             &55.00 &52.0 &34.1 &72.4 &69.2 &35.7 &62.2 &57.2     \\
 + \textbf{Ours}               &            &8.55 &59.0 &66.1 &73.2 &69.4 &36.5 &64.0 &61.4        \\
\midrule
+ LTMP           & \multirow{2}{*}{20\%}             &466.40 &38.4 &27.7 &63.5 &64.7 &33.1 &63.8 &48.5      \\
+ \textbf{Ours}               &             &17.96 &49.1 &64.7 &68.2 &69.4 &37.5 &63.1 &58.7       \\
\midrule
+ LTMP           & \multirow{2}{*}{30\%}             &4670.71 &22.3 &24.9 &58.9 &54.0 &28.3 &59.2 &41.3      \\
 + \textbf{Ours}               &            &42.61 &38.3 &59.4 &61.2 &68.4 &37.3 &63.9 &54.7       \\
\bottomrule
\end{tabular}
}
\caption{Additional results of post-training performance on Mamba-2-2.7B. We compare with LTMP and evaluate them on six benchmarks under 10\%, 20\%, and 30\% FLOPS reduction.}
\label{tab:our_ltmp}
\end{table*}
The throughput of token generation for Mamba-1.4B and Mamba-2-1.3B using the proposed method are also collected under the same configuration in Section \ref{sec:efficient_results}, as illustrated in Figure~\ref{fig:throughput-small}.
With our optimization, the throughput can be improved by $1.08\times$, $1.15\times$, and $1.26\times$ for Mamba-1.4B, and $1.10\times$, $1.19\times$, and $1.35\times$ for Mamba-2-1.3B, when reducing 10\%, 20\%, and 30\% FLOPS, respectively.

\subsection{More Results}
We compared our method with LTMP~\cite{bonnaerens2023learned}, a simple token pruning and merging method designed for Vision Transformer.
Our method outperforms LTMP in six benchmarks under same FLOPS reduction by a large margin, as shown in Table~\ref{tab:our_ltmp}.
The results emphasizing that the simple combination of token pruning and merging from Transformer is inadequate for SSMs.

\end{document}